\documentclass[11pt]{article}

\usepackage[preprint]{acl}
\usepackage{times}
\usepackage{latexsym}
\usepackage{amsmath}
\usepackage{amssymb}
\usepackage{graphicx}
\usepackage{booktabs}
\usepackage{multirow}
\usepackage{xcolor}
\usepackage{subcaption}
\usepackage{natbib}
\usepackage{hyperref}
\usepackage{times}
\usepackage{lipsum}
\usepackage{latexsym}
\usepackage[T1]{fontenc}

\usepackage[utf8]{inputenc}

\usepackage{microtype}

\usepackage{inconsolata}

\usepackage{graphicx}

\newcommand{\frs}{S_{\mathrm{FR}}}
\newcommand{\Df}{\Delta_{\mathrm{forget}}}
\newcommand{\Dr}{\Delta_{\mathrm{retain}}}
\newcommand{\fenc}{f_{\mathrm{enc}}}
\newcommand{\fdec}{f_{\mathrm{dec}}}

\title{Understanding Sparse Autoencoder Dictionary Size for Unlearning and Mechanistic Interpretability in LLMs}

\author{Amartya Hatua \\
  Fidelity Investments \\
  \texttt{amartyahatua@gmail.com} \\\And
  Filip Cano \\
  Institute of Science and Technology Austria \\
  \texttt{filip.cano@ist.ac.at}\\\AND
  Badri Narayan Subudhi \\
   Indian Institute of Technology Jammu \\
  \texttt{subudhi.badri@gmail.com} 
  }

\begin{document}
\maketitle

\begin{abstract}
Machine unlearning aims to remove targeted knowledge from language models while preserving retained capabilities. This paper studies how sparse autoencoder (SAE) design choices of dictionary size and layer placement affect SAE-guided unlearning and mechanistic interpretability. 
With ``task of ficticious unlearning for LLMs'' (TOFU) as our benchmark, we train SAEs across layers and dictionary sizes from 4K to 64K for GPT-2 Small, Medium, Large, and Llama 3.2 1B, then suppress features most specific to forget examples during inference. Results\footnote{All the evaluation source code and data can be accessed in \url{https://github.com/amartyahatua/EACL_2027}} show that dictionary size has a scale-dependent effect: larger dictionaries substantially improve forget–retain selectivity in larger models, but provide little or no benefit for smaller models. This pattern aligns with interpretability measurements, where larger dictionaries reduce feature overlap in large models. 
Layer choice also affects unlearning, but no consistent cross-model pattern arises.
\end{abstract}

\begin{figure*}[t] 
    \centering
    \includegraphics[width=\textwidth]{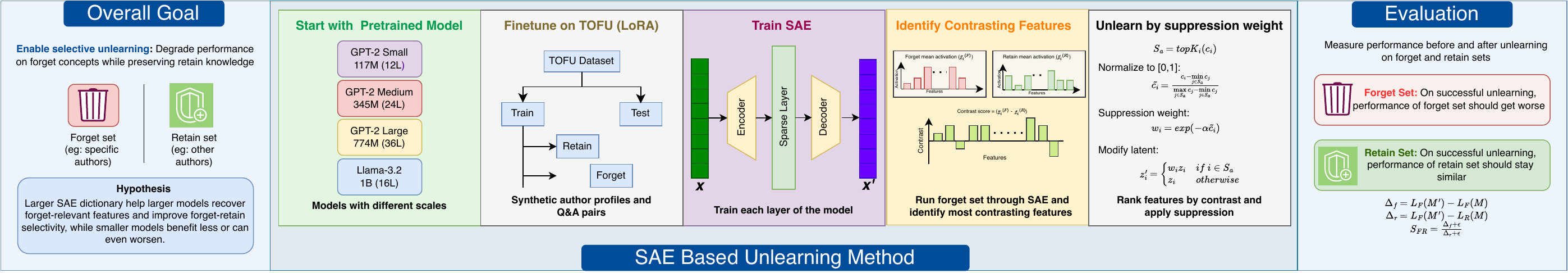}
    \caption{SAE-Based Unlearning}
    \label{fig:wide_image}
\end{figure*}

\section{Introduction}

Large language models (LLMs) can memorize and reproduce training information that may later need to be removed, because of privacy or correctness concerns. Machine unlearning aims to remove targeted information while preserving general model utility~\citep{cao2015towards}. 
Effective unlearning should degrade the performance only on the topics designated to forget, while causing little to no degradation of the model on the information to retain. Existing unlearning methods often optimize this tradeoff directly in parameter space, but they can be difficult to interpret mechanistically~\citep{ChenY23}.

Sparse autoencoders (SAEs) have recently gained traction as a method for mechanistic interpretability in LLMs~\cite{cunningham2024sparse,bricken2023monosemanticity}.
By training an overcomplete autoencoder on hidden activations and enforcing sparsity, SAEs can learn latent features that are often more interpretable than individual neurons. 
SAE-based unlearning is based on detecting which latent features correspond to concepts, entities, facts, or behaviours that should be removed, by identifying the most activated latent features when inferring answers in the forget set, and damping or knocking out the corresponding latent features during inference.

While the use of SAEs for mechanistic interpretability and unlearning is quite extended, there are two design choices that have received little attention: layer and dictionary size. 
These are generally treated as either unimportant, or as hyperparameters to fine-tune on a per application basis.
 Fitting an SAE at a layer that does not influence a concept or task may yield misleading explanations and ineffective unlearning. A dictionary that is too small may merge unrelated concepts into polysemantic features, whereas an overly large dictionary may waste capacity or leave many features underused.

In this work, we empirically study how SAE layer positioning and dictionary size affect SAE-based unlearning and interpretability. We use the TOFU benchmark~\citep{maini2024tofu}, which consists of synthetic author profiles designed to test whether a model can forget specific individuals while retaining other knowledge.
Our main finding is that dictionary size has a model-dependent effect. In large models, increasing dictionary size improves unlearning selectivity noticeably. However, in small models, 
dictionary size has a negligible effect on unlearning.
We structure our investigation around two research questions:\\
\textbf{(RQ1):} Under a fixed SAE training and intervention procedure, how do dictionary size and layer placement affect forget--retain selectivity?\\ 
\textbf{(RQ2):} Given that SAE size changes selectivity, is that change associated with better separation between forget- and retain-relevant SAE features? \\

\noindent\textbf{Contributions.}
\begin{itemize}
    \item We report a systematic study of SAE-based unlearning across model scales, layers, and dictionary sizes.
    \item We show that larger dictionaries improve SAE-guided unlearning most clearly for larger models, while smaller models show little or no monotonic benefit.
    \item We compare this scaling behavior with the overlap between features active on forget and retain examples, connecting the reported phenomenon to feature polysemanticity, and feature separation in the SAE.
    \item We show that there is also considerable variability accross layer choice, but no specific pattern arises for different models.
\end{itemize}


\section{Methodology}

\subsection{General Setup}
We use TOFU~\citep{maini2024tofu} as our benchmark for unlearning.
TOFU consists of synthetic author profiles paired with question-answer examples. 
Models are first fine-tuned with the TOFU dataset using Low-Rank Adaptation (LoRA)~\citep{hu2022lora}.
For each model, we evaluate unlearning 
by contrasting performance differences before and after unlearning.
We use the GPT-2~\citep{radford2019language} family and Llama 3.2 1B~\cite{llama} as our subject models, summarized in Table~\ref{tab:model-sizes}.

\begin{table}[b]
    \centering
    \footnotesize
    \begin{tabular}{lccc}
    \toprule
         Model & No. Params. & No. Layers & Layer size  \\ \midrule
         GPT2-Small & 117M & 12 & 768\\
         GPT2-Med & 345M & 24 & 1024\\
         GPT2-Large & 774M & 36 & 1280\\
         Llama & 1B & 16 & 2048\\
         \bottomrule
    \end{tabular}
    \caption{Model sizes}
    \label{tab:model-sizes}
\end{table}



For each model, we train SAEs on all layers, with dictionary sizes $D \in \{4\mathrm{K}, 8\mathrm{K}, 16\mathrm{K}, 32\mathrm{K}, 64\mathrm{K}\}$.
These correspond to expansion factors $(D/d)$ between $2.0\times$ and $85.3\times$, as we detail in 
Table~\ref{tab:expansion-factors}. 

\begin{table}[t]
\centering
\scriptsize
\caption{Dictionary expansion factors \(D/d\) for each subject model.}
\label{tab:expansion-factors}
\resizebox{\columnwidth}{!}{%
\begin{tabular}{lrrrrr}
\toprule
Model & 4K & 8K & 16K & 32K & 64K \\
\midrule
GPT-2 Small  & $5.3\times$ & $10.7\times$ & $21.3\times$ & $42.7\times$ & $85.3\times$ \\
GPT-2 Medium & $4.0\times$ & $8.0\times$  & $16.0\times$ & $32.0\times$ & $64.0\times$ \\
GPT-2 Large  & $3.2\times$ & $6.4\times$  & $12.8\times$ & $25.6\times$ & $51.2\times$ \\
Llama 1B     & $2.0\times$ & $4.0\times$  & $8.0\times$  & $16.0\times$ & $32.0\times$ \\
\bottomrule
\end{tabular}}
\end{table}

All experiments were performed on an Nvidia A100 80 GB, and consumed a total of 100 hours of compute time.

\subsection{Sparse Autoencoders}

For a model \(M\) and layer \(\ell\), let \(h_{\ell}(x) \in \mathbb{R}^{d}\) denote the hidden activation of input \(x\) at that layer. We train a sparse autoencoder with encoder \(\fenc\) and decoder \(\fdec\) to reconstruct these activations through a sparse latent representation:
\begin{align}
    z(x) &= \fenc(h_{\ell}(x)) \in \mathbb{R}^{D} \\
    \hat{h}_{\ell}(x) &= \fdec(z(x)) \in \mathbb{R}^{d}
\end{align}
The dictionary size \(D\) is larger than the hidden dimension \(d\), giving the SAE an overcomplete set of latent features. Sparsity is enforced using the same training objective and sparsity protocol across dictionary sizes, so that changes in unlearning behavior can be attributed primarily to the dictionary-size sweep. 

\subsection{SAE architecture and training setup.}

For each model, layer, and dictionary size, we train a separate SAE. To isolate the effect of dictionary size, all SAEs use the same standard ReLU architecture rather than TopK or JumpReLU variants. Given a residual-stream activation \(h_{\ell}(x)\in\mathbb{R}^{d}\), the encoder and decoder are defined as 

\begin{align} z(x) &= \mathrm{ReLU}\!\left( W_{\mathrm{enc}}h_{\ell}(x)+b_{\mathrm{enc}} \right) \\ \hat{h}_{\ell}(x) &= W_{\mathrm{dec}}z(x) \end{align}

The SAE is trained to accurately reconstruct the original activation while encouraging sparse latent representations. The training objective is:

\begin{equation} \begin{split} \mathcal{L}_{\mathrm{SAE}} &= \frac{1}{N}\sum_{n=1}^{N} \left\| h_{\ell}(x_n)-\hat{h}_{\ell}(x_n)\right\|_2^2 \\ &\quad + \lambda \|z(x_n)\|_1 \end{split} \end{equation}

where the first term is the reconstruction loss, the second term encourages sparsity in the latent features, and \(\lambda\) controls the sparsity-reconstruction trade-off. Unless otherwise specified, the same training hyperparameters and sparsity settings are used across all dictionary sizes.

Complete SAE architectural and LoRA fine-tuning settings are provided in Appendix~\ref{sec:more_experiments}. These hyperparameters were kept fixed within each model family and varied only by dictionary size. To ensure the validity of the learned representations, we conducted null-test quality-control checks for every model-dictionary configuration. Aggregate pass/fail statistics are reported in Appendix Table~\ref{tab:sae_null_test}. The high overall pass rate (96.8\%) indicates that the learned SAEs satisfy the required quality criteria in the vast majority of cases.

\subsection{SAE-Guided Selective Unlearning}
\label{sec:sae-unlearning}

We perform unlearning by suppressing SAE features that are most specific to the forget data. For each target author \(a\), we first compute the mean SAE activation on the forget set and retain set:
\begin{align}
    \bar{\mathbf{z}}^{(F)}_a &= \mathbb{E}_{x \in \mathcal{D}^{(F)}_a}[\fenc(h_{\ell}(x))] \\
    \bar{\mathbf{z}}^{(R)}_a &= \mathbb{E}_{x \in \mathcal{D}^{(R)}_a}[\fenc(h_{\ell}(x))]
\end{align}
These vectors summarize how strongly each SAE feature responds to forget and retain examples.

We then define a contrast score for each feature:
\begin{equation}
    c_i = \bar{z}^{(F)}_{a,i} - \bar{z}^{(R)}_{a,i}
    \label{eq:contrast}
\end{equation}
Features with large positive contrast activate more strongly on the forget author than on retained authors, and are therefore candidates for intervention. We rank features by \(c_i\) and select the top \(K\) features as the intervention set \(\mathcal{S}_a\).
We study interventions with $K\in \{50,100,200,500\}$.

During inference, we suppress the selected features at layer \(\ell\). Rather than applying a uniform ablation to all selected features, we scale each selected feature according to its contrast score. We normalize the selected contrast scores to \([0,1]\):
\begin{equation}
\tilde{c}_i =
\frac{c_i - \min_{j \in \mathcal{S}_a} c_j}
     {\max_{j \in \mathcal{S}_a} c_j - \min_{j \in \mathcal{S}_a} c_j}
\end{equation}
and assign each selected feature a suppression weight
    $w_i = \exp(-\alpha \tilde{c}_i)$,
where \(\alpha\) controls the strength of the intervention. 
In our experiments, we use $\alpha = $ 4.
The modified latent representation is
\begin{equation}
z'_i =
\begin{cases}
  w_i z_i & \text{if } i \in \mathcal{S}_a, \\
  z_i     & \text{otherwise}.
\end{cases}
\label{eq:ablation}
\end{equation}
The decoder maps \(z'\) back into the model's hidden space, producing the intervened representation used for the rest of the computation. This procedure leaves all non-selected SAE features unchanged, and therefore directly tests whether the selected sparse features provide a selective handle on the forget information.

\subsection{Evaluation Metrics}

We measure unlearning by comparing model loss before and after intervention. Let \(L_A(M)\) denote the language-model loss of model \(M\) on dataset \(A\), where \(A\) is either the forget set \(F\) or retain set \(R\). For the original model \(M\) and intervened model \(M'\), we define
\begin{align}
    \Df &= L_F(M') - L_F(M) \\
    \Dr &= L_R(M') - L_R(M)
\end{align}
A useful unlearning intervention  should produce a large \(\Df\), indicating that forget-set performance has degraded, and a small \(\Dr\), indicating that retain-set performance is preserved.
We summarize this trade-off using selectivity:
\begin{equation}
    \frs = \frac{\Df + \epsilon}{\Dr + \epsilon}
\end{equation}
where \(\epsilon = 10^{-5}\) is included for numerical stability. Higher \(\frs\) indicates more selective unlearning, and \(\frs > 1\) means that the intervention harms forget examples more than retain examples. 

\subsection{Feature Purity and Polysemanticity}
\label{sec:purity}

To understand why dictionary size affects unlearning differently across model scales, we measure how well SAE features separate forget and retain data. The guiding intuition is that selective unlearning is easier when forget examples activate features that are not also strongly used by retain examples. Conversely, if forget and retain examples activate many of the same features, then those features are more polysemantic with respect to the unlearning task, and suppressing them is more likely to damage retained knowledge.
For each forget author \(a\), we define the active feature sets $S_F(\tau)$ and $S_R(\tau)$:
\begin{align}
S_{F/R}(\tau) &= \{\, i : \bar{z}^{(F/R)}_{a,i} > \tau \,\}
\end{align}
We then compute their Jaccard overlap:
\begin{equation}
J(\tau) =
\frac{\lvert S_F(\tau) \cap S_R(\tau) \rvert}
     {\lvert S_F(\tau) \cup S_R(\tau) \rvert}
\label{eq:jaccard}
\end{equation}
Lower overlap indicates cleaner separation between forget and retain features, while higher overlap indicates that the same features are shared across both sets. We interpret high overlap as an indication of polysemanticity for the purpose of unlearning.

Because many SAE features may correspond to generic information unrelated to the target author, 
we first compute the overlap among the features most relevant to the intervention.
For each author, let \(T_a\) contain the top 2,048 features ranked by contrast \(c_i\). 
We compute Jaccard overlap between \(S_F(0)\cap T_a\) and \(S_R(0)\cap T_a\), after removing features with \(\bar z^{(F)}_{a,i}+\bar z^{(R)}_{a,i}<10^{-6}\), and average over the 19 forget authors. Ranking only by \(\bar z^{(F)}_{a,i}\) would also select generic features active on both sets, so it would not help us quantify selective intervention.

The fixed cutoff covers 50\% of a 4K dictionary and 3.125\% of a 64K dictionary, and its coverage also depends on how many features are alive. We therefore repeat the calculation over all alive features, using the same activity filter and $\tau=0$ without the top-2,048 cutoff.


\begin{table}[b]
\scriptsize
\centering
\caption{Mean forget--retain selectivity with 95\% confidence intervals.}
\label{tab:selectivity_dict_ci}
\begin{tabular}{lllll}
\toprule
$D$ &      GPT2 Small &       GPT2 Med. &      GPT2 Large &           Llama \\
\midrule
 4K & 2.34 $\pm$ 0.08 & 2.55 $\pm$ 0.08 & 1.08 $\pm$ 0.03 & 3.08 $\pm$ 0.13 \\
 8K & 2.27 $\pm$ 0.08 & 2.60 $\pm$ 0.08 & 1.10 $\pm$ 0.03 & 3.57 $\pm$ 0.15 \\
16K & 2.17 $\pm$ 0.10 & 2.62 $\pm$ 0.19 & 4.32 $\pm$ 0.28 & 4.35 $\pm$ 0.38 \\
32K & 2.06 $\pm$ 0.11 & 2.92 $\pm$ 0.19 & 5.18 $\pm$ 0.29 & 4.63 $\pm$ 0.29 \\
64K & 2.03 $\pm$ 0.12 & 3.48 $\pm$ 0.86 & 5.52 $\pm$ 0.25 & 6.06 $\pm$ 0.38 \\
\bottomrule
\end{tabular}
\end{table}

\section{Results}
\begin{figure}[t]
    \centering
    \includegraphics[width=\linewidth]{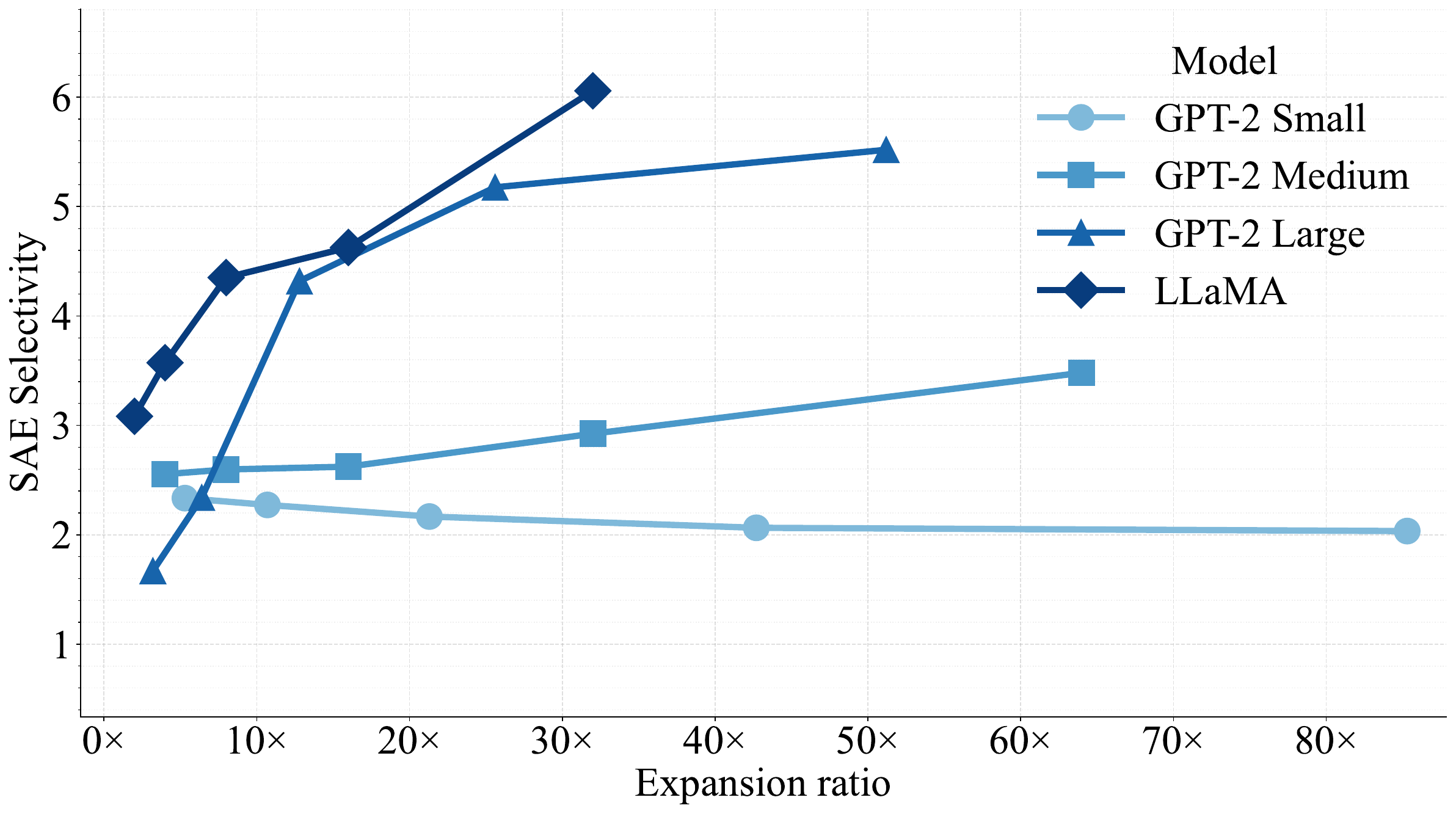}
        \caption{Selectivity against dictionary expansion factor.}
        \label{fig:normalized-size}
\end{figure}

\paragraph*{RQ1: SAE Size and Layer Placement.}
In Table~\ref{tab:selectivity_dict_ci} we report the unlearning performance, in terms of selectivity, for all four models, averaged over all layers. We observe the tendency that selectivity increases notably with dictionary size for the larger model, while the effect on medium and small sized models is between moderate and null.
In Figure~\ref{fig:normalized-size}, we plot these observations against the expansion factor $D/d$, showing a strong upward trend for the larger models, a mild upward trend for GPT-Medium, and an almost flat (slightly downward) trend for GPT-Small.



\begin{figure}[t]
\centering
    \includegraphics[width=\linewidth]{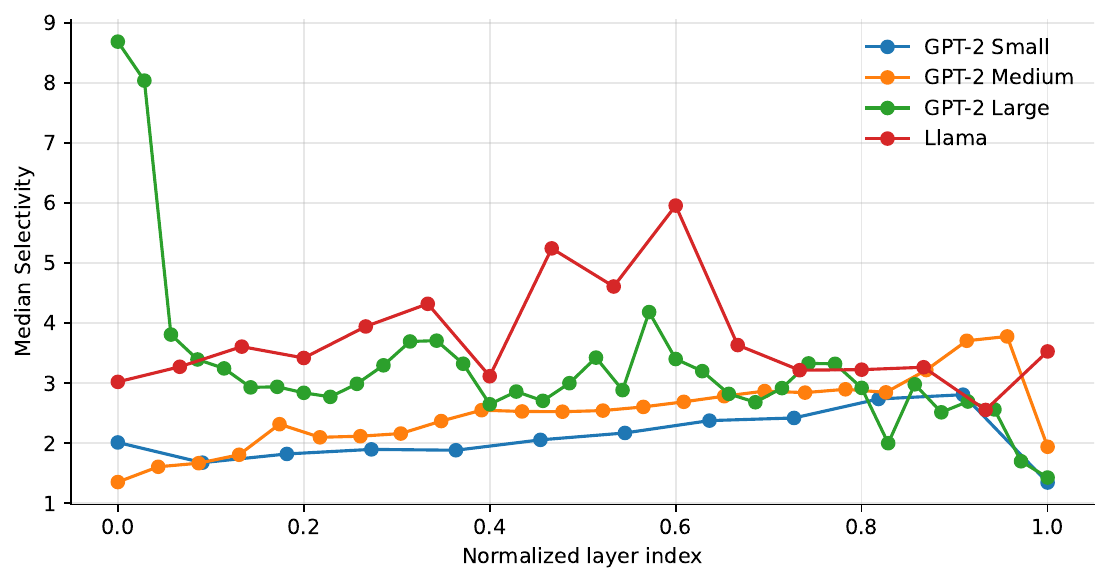}
        \caption{Median selectivity by normalized layer; each point pools authors and dictionary sizes.}
        \label{fig:selectivity-layer}
\end{figure}

In Figure~\ref{fig:selectivity-layer} we show the median selectivity accross model layers.
As for layer selection, unlearning varies substantially accross layers, but there is no apparent trend.
 We illustrate these results in Figure~\ref{fig:selectivity-layer}, where we plot for each model and each layer, the median selectivity accross authors and SAE dictionary sizes. 
 Thus, there is no a priori knowledge in which layer to choose to implement forgetting.

\paragraph*{RQ2: Feature Separation.}
Table~\ref{tab:jaccard_dict} reports overlap for both sets of features. Among the top 2,048 features, the overlap falls strongly for GPT-2 Medium (0.9727 to 0.4554) and GPT-2 Large (0.8999 to 0.4677), while GPT-2 Small approaches complete overlap. The trend is weaker in Llama, where the downward pattern exists in general, but it is sometimes non-monotonic. For example, from 4K to 16K its top-2,048 overlap rises from 0.9089 to 0.9526 while $S_{FR}$ also rises, from 3.08 to 4.35. Using all alive features, overlap stays roughly constant for GPT-2 Small and decreases overall for the other three models.

\section{Discussion}
\paragraph*{RQ1.} Our results suggest that larger dictionaries are not a universally beneficial parameter, and that larger models tend to benefit from larger dictionaries, even when normalising by model layer size, while medium and small models tend to be less affected. 
They also show that SAE layer placement follows no general trend, suggesting that layer choice is a model and application dependent hyperparameter, to optimise on its own.

\paragraph*{RQ2.} The overlap results are consistent with the hypothesis that wider SAEs help when they separate forget-relevant from retain-relevant activity. This interpretation aligns with the motivation that sparse dictionaries can expose features more cleanly than individual neurons~\citep{elhage2022toy,bricken2023monosemanticity,cunningham2024sparse}.
Smaller models do not benefit from larger dictionaries, suggesting that wider SAEs help only when the model contains separable semantic structure.



\begin{table}
\footnotesize
\centering
\caption{Jaccard overlap by SAE dictionary size for the top 2,048 features by contrast and for all alive features.}
\label{tab:jaccard_dict}
\begin{tabular}{lcccc}
\toprule
$D$ & GPT2 S. & GPT2 M. & GPT2 L. &  Llama \\
\midrule
\multicolumn{5}{l}{ {\emph{Top 2,048 by contrast}}} \\
 {4K} &  {0.885} &  {0.972} &  {0.899} &  {0.909} \\
 {8K} &  {0.905} &  {0.917} &  {0.804} &  {0.949} \\
 {16K} &  {0.999} &  {0.654} &  {0.694} &  {0.953} \\
 {32K} &  {0.999} &  {0.473} &  {0.586} &  {0.793} \\
 {64K} &  {0.995} &  {0.455} &  {0.468} &  {0.739} \\
\midrule
\multicolumn{5}{l}{\emph{All alive features}} \\
 4K &     0.960 &    0.902 &     0.902    & 0.944 \\
 8K &     0.964 &    0.891 &     0.891 & 0.837 \\
16K &     0.965 &    0.844 &     0.871 & 0.343 \\
32K &     0.968 &    0.809 &     0.826 & 0.360 \\
64K &     0.964 &    0.805 &     0.801 & 0.331 \\
\bottomrule
\end{tabular}
\end{table}
\section{Related Work}

Machine unlearning was originally framed as removing the influence of data without full retraining \citep{cao2015towards}, and has gained traction in the recent years, both in general unlearning methods~\citep{ginart2019making,bourtoule2021machine,guo2020certified,neel2021descent,warnecke2023machine},
as well as LLM-specific methods~\citep{jang2023knowledge,maini2024tofu,yao2024machine,yao2024large,li2024wmdp,pawelczyk2024incontext,zhang2024negative,shi2025muse}. 

SAEs were introduced as a method for mechanistic interpretability~\citep{geva2021transformer,dai2022knowledge,meng2022locating,meng2023mass,liu2024targeted},
and have become a strong and interpretable method for unlearning in LLMs:~\citep{farrell2024applying} study SAE feature interventions for removing hazardous knowledge, \citep{khoriaty2025dont,muhamed2025saes}  use conditional SAE clamping for dynamic guardrails, and \citep{wang2025modelunlearning} use SAE-selected subspaces to guide persistent parameter updates for unlearning.
Our contribution is analytical: we isolate how SAE layer and dictionary size affect the selectivity of SAE-guided interventions. More broadly, our results build on work showing that sparse dictionaries can recover interpretable features, but that feature quality has dependencies on scale, task, and evaluation procedure that are not yet well-understood~\citep{bricken2023monosemanticity,cunningham2024sparse,gao2025scaling,lieberum2024gemma,makelov2025principled,marks2025sparse,leask2025canonical}.

\section{Conclusion \& Future Work}

We studied how the choice of layer and dictionary size affect SAE-guided unlearning and interpretability.
Our main observation is that unlearning improves with dictionary size only for large enough models, and that this improvement seems to be due to the increase in feature separation achieved by larger dictionary sizes in models that are not too polysemantic.

In future work, these results should be tested in the future in broader benchmarks, such as MUSE, WMDP, and OpenUnlearning~\cite{shi2025muse,li2024wmdp,dorna2025openunlearning}, and a broader family of models, to ascertain how prevalent is the persisten phenomenon.
This would also lead to a more principled understanding of how polysemanticity affects unlearning. 
Finally, another exciting research direction is to develop adaptive size and layer selection methods, to find optimal values of the hyperparameters for a concrete model and task.

\section*{Limitations}
We choose TOFU~\cite{maini2024tofu} deliberately for isolating forgetting effects.
Because its author profiles are synthetic, and TOFU was released after the training of the studied models, we can ensure that the target knowledge is introduced only through controlled fine-tuning, and the knowledge to be forgotten is not entangled with uncontrolled pre-existing knowledge about real entities in the training corpus. This allows us to best isolate the effect of SAE dictionary size and layer placement.
However, real unlearning requests may involve noisier boundaries, paraphrases, partial overlap with retained knowledge, and adversarial extraction attempts.
Extending our work to other benchmarks containing real-world data, such as MUSE~\cite{shi2025muse}, WMDP~\cite{li2024wmdp}, or OpenUnlearning~\cite{dorna2025openunlearning} would be very valuable and would help verify the trends found in this paper to broader and more realistic settings.

Our model coverage is representative of the open source state of the art, but also limited, and further experiments with more models would also help corroborate the studied effect.

A further limitation is that the SAE-based unlearning studied here is pure feature ablation. We suppress selected SAE features during inference, but we do not rewrite the original model parameters. Thus, the intervention only applies while the SAE module is inserted into the forward pass. A stronger form of persistent unlearning would require committing the modified computation back into a layer with the same architecture and input-output interface as the original model. We leave this step outside the scope of this work, since our goal is to study feature separation and selectivity rather than to build a full deployment-ready unlearning procedure.

\section*{Ethical Considerations}

Machine unlearning can support important goals such as privacy protection, correction of outdated or erroneous information, and compliance with legitimate deletion requests. SAE-based methods may be useful in this setting because they provide an interpretable handle on which internal features are being suppressed.

However, the interventions studied here should not be presented as certified deletion. They do not permanently remove information from the model, and benchmark-level loss changes do not rule out recovery through adversarial prompting or other extraction methods. In sensitive deployments, such methods would require stronger evaluation, auditing, and clear reporting of failure modes.

Selective unlearning also creates misuse risks. Similar techniques could be used to suppress inconvenient facts, weaken safety behavior, or remove information unevenly across groups or domains. Because SAE features may be shared across related concepts, ablation can also cause collateral damage to retained knowledge. These risks should be considered before applying SAE-guided unlearning outside controlled benchmarks.

\section*{Acknowledgments}
This work was supported by the European Research Council under Grant No.: ERC-2020-AdG 101020093 (VAMOS). This work was also supported by the ISTA Responsible AI Program, made possible through the support of Garrett Camp and the Camp Foundation.

\bibliography{references}

\clearpage

\appendix





\section{Additional Experimental Details Evidence}
\label{sec:more_experiments}
In this appendix we include details about the empirical evaluation's setup and results that do not fit within the main paper.

\subsection{SAE Training Settings}
\label{sec:sae_training}
For each dictionary size $D\in\{4\mathrm{K},8\mathrm{K},16\mathrm{K},32\mathrm{K},64\mathrm{K}\}$, we train ReLU SAEs on the TOFU \texttt{retain90} split using Adam with a batch size of 256. The objective combines mean squared reconstruction error with an $\ell_1$ sparsity penalty $\lambda_D\,\operatorname{mean}(|z|)$ on the latent activations $z$. For GPT-2 Small, Medium, and Large (hidden dimensions 768, 1024, and 1280, respectively), we train for eight epochs with a learning rate of $3\times10^{-4}$ and $\lambda_D=5\times10^{-5}(4096/D)$. For Llama-3.2-1B (hidden dimension 2048), we train for ten epochs with both the learning rate and $\lambda_D$ fixed at $10^{-3}$.

To evaluate success in training, we perform $95\%$ confidence quality control tests, and report the results in Table~\ref{tab:sae_null_test}.

For fine-tuning, we use LoRA targeted on modules 
\texttt{c\_attn,c\_proj,c\_fc} for the GPT-2 family, and 
\texttt{q,k,v,o,gate,up,down proj.} for Llama-3.2. 
The full hyperparameter selection is reported in Table~\ref{tab:lora_config}.


\subsection{Disambiguated Selectivity Values}
We define selectivity ($S_{FR}$) as a quotient between the differences of forget and retain loss. This is useful as a compact summary of the forget-retain tradeoff. A large ratio can result either from a substantial increase in forgetting, or from a very small increase in retain loss. Both indicate a somewhat successful unlearning step, but in different directions.
No single scalar can completely resolves this ambiguity. 

To show the scale of the different factors, we report in Table~\ref{tab:raw_delta_other} for all dictionary sizes the averaged values of $\Delta_{\mathrm{forget}}$ and $\Delta_{\mathrm{retain}}$, as a disambiguted version of $(S_{FR})$.

\begin{table}[b]
    \centering
    \footnotesize
    \caption{Forget and retain changes computed from median perplexities. 
    Values are rounded to one decimal place.}
    \label{tab:raw_delta_other}
    \setlength{\tabcolsep}{3pt}
    \begin{tabular}{llrrrr}
    \toprule
    $D$ & & GPT2 Small & GPT2 Med. & GPT2 Large & Llama \\
    \midrule
4K & $\Df$ & 94.3 & 17.5 & 20.6 & 8.2 \\
 & $\Dr$ & 42.1 & 7.0 & 12.7 & 3.0 \\
 & $S_{FR}$ & 2.2 & 2.5 & 1.6 & 2.7 \\
\addlinespace[2pt]
8K & $\Df$ & 55.0 & 35.5 & 7.4 & 14.2 \\
 & $\Dr$ & 24.9 & 13.9 & 3.7 & 4.2 \\
 & $S_{FR}$ & 2.2 & 2.6 & 2.0 & 3.4 \\
\addlinespace[2pt]
16K & $\Df$ & 30.6 & 9.8 & 193.8 & 34.3 \\
 & $\Dr$ & 14.5 & 3.8 & 54.1 & 9.5 \\
 & $S_{FR}$ & 2.1 & 2.6 & 3.6 & 3.6 \\
\addlinespace[2pt]
32K & $\Df$ & 24.8 & 5.4 & 131.2 & 67.3 \\
 & $\Dr$ & 12.9 & 2.0 & 28.2 & 15.9 \\
 & $S_{FR}$ & 1.9 & 2.7 & 4.7 & 4.2 \\
 \addlinespace[2pt]
 64K & $\Df$ & 18.3 & 3.4 & 108.7 & 132.7 \\
 & $\Dr$ & 10.3 & 1.4 & 21.7 & 23.8 \\
 & $S_{FR}$ & 1.8 & 2.4 & 5.0 & 5.6 \\
    \bottomrule
    \end{tabular}
\end{table}

Note that these ratios differ from the corresponding entries in Table~\ref{tab:selectivity_dict_ci} because Table~\ref{tab:selectivity_dict_ci} reports the mean of the per-experiment selectivity scores, whereas this table first averages the losses before and after separately and then computes the differences and their ratio.

\begin{table}[b]
\centering
\small
\caption{SAE null-test quality control across models.
A configuration passes when $\mathrm{ICC}<0.05$.
Each configuration corresponds to a dictionary-size/layer pair.}
\label{tab:sae_null_test}
\begin{tabular}{lrrrr}
\toprule
\textbf{Model} &
\textbf{Tested} &
\textbf{Pass} &
\textbf{Fail} &
\textbf{Pass rate} \\
\midrule
GPT-2 Small      & 59  & 59  & 0  & 100.0\% \\
GPT-2 Medium     & 119 & 117 & 2  & 98.3\% \\
GPT-2 Large      & 180 & 178 & 2  & 98.9\% \\
Llama-3.2-1B     & 80  & 70  & 10 & 87.5\% \\
\midrule
\textbf{Overall} & \textbf{438} & \textbf{424} &
\textbf{14} & \textbf{96.8\%} \\
\bottomrule
\end{tabular}
\end{table}

\begin{table}[t]
\centering
\footnotesize
\caption{LoRA fine-tuning configuration. The same configuration is used
for GPT-2 Small, Medium, and Large.}
\label{tab:lora_config}
\begin{tabular}{lcc}
\toprule
\textbf{Hyperparameter} &
\textbf{GPT-2} &
\textbf{Llama-3.2-1B} \\
\midrule
LoRA rank $r$              & 32 & 16 \\
LoRA $\alpha$              & 64 & 16 \\
LoRA dropout               & 0.05 & 0.0 \\
LoRA scaling $\alpha/r$    & 2.0 & 1.0 \\
Training epochs            & 15 & 10 \\
Per-device batch size      & 8 & 4 \\
Gradient accumulation      & 4 & 4 \\
Effective batch size       & 32 & 16 \\
Learning rate              & $2\times10^{-4}$ & $2\times10^{-4}$ \\
Weight decay               & 0.01 & 0.01 \\
LR scheduler               & Cosine & Cosine \\
Warm-up                    & 0.1 ratio & 20 steps \\
Max. seq. length    & 512 & 512 \\
\bottomrule
\end{tabular}
\end{table}

\end{document}